\documentclass[11pt,a4paper]{article}
\usepackage[colorlinks,linkcolor=blue]{hyperref}
\usepackage[hyperref]{acl2019}
\usepackage{times}
\usepackage{latexsym}
\usepackage{multirow} 
\usepackage{multicol}
\usepackage{url}

\aclfinalcopy 
\def\aclpaperid{2066} 

\usepackage{graphicx}
\usepackage{amsmath}

\usepackage{subfig}

\usepackage{latexsym}
\usepackage{array}
\usepackage{xcolor}
\usepackage{bm}
\usepackage{multirow}
\usepackage{booktabs}
\usepackage{amsfonts} 
\usepackage[ruled,vlined]{algorithm2e} 

\newcommand{\tabincell}[2]{
\begin{tabular}{@{}#1@{}}#2\end{tabular}
}

\title{Learning Representation Mapping for Relation Detection in \\ Knowledge Base Question Answering}

\author{Peng Wu$^{1,2}$, Shujian Huang$^{1,2}$
, Rongxiang Weng$^{1,2}$, Zaixiang Zheng$^{1,2}$\textbf{,}\\ 
\textbf{Jianbing Zhang}$^{1,2}$\textbf{,} \textbf{Xiaohui Yan}$^{3}$ \and \textbf{Jiajun Chen}$^{1,2}$\\
  $^{1}$National Key Laboratory for Novel Software Technology, Nanjing, China \\
  $^{2}$Nanjing University, Nanjing, China \\
   $^{3}$Poisson Lab, Huawei Technologies, Beijing, China \\
    {\tt \{wup, wengrx, zhengzx\}@nlp.nju.edu.cn} \\
    {\tt \{huangsj, zjb, chenjj\}@nju.edu.cn}\\
    {\tt yanxiaohui2@huawei.com} 
         }

\date{}

\begin{document}

\maketitle

\begin{abstract}
  Relation detection is a core step in many natural language process applications including knowledge base question answering. Previous efforts show that single-fact questions could be answered with high accuracy. However, one critical problem is that current approaches only get high accuracy for questions whose relations have been seen in the training data. But for unseen relations, the performance will drop rapidly. The main reason for this problem is that the representations for unseen relations are missing. In this paper, we propose a simple mapping method, named representation adapter, to learn the representation mapping for both seen and unseen relations based on previously learned relation embedding. We employ the adversarial objective and the reconstruction objective to improve the mapping performance. We re-organize the popular SimpleQuestion dataset to reveal and evaluate the problem of detecting unseen relations. Experiments show that our method can greatly improve the performance of unseen relations while the performance for those seen part is kept comparable to the state-of-the-art.\footnote{Our code and data are available at  \url{https://github.com/wudapeng268/KBQA-Adapter}.}
  
\end{abstract}

\section{Introduction}
The task of Knowledge Base Question Answering (KBQA) has been well developed in recent years~\cite{berant2013semantic,bordes2014question,yao2014information}. It answers questions using an open-domain knowledge base, such as Freebase \cite{bollacker2008freebase}, DBpedia \cite{lehmann2015dbpedia} or NELL \cite{carlson2010toward}. The knowledge base usually contains a large set of triples. Each triple is in the form of $\langle$\textit{subject, relation, object}$\rangle$, indicating the \textit{relation} between the \textit{subject entity} and the \textit{object entity}.

\begin{figure}[t]
\centering
{
\includegraphics[width=7.5cm]{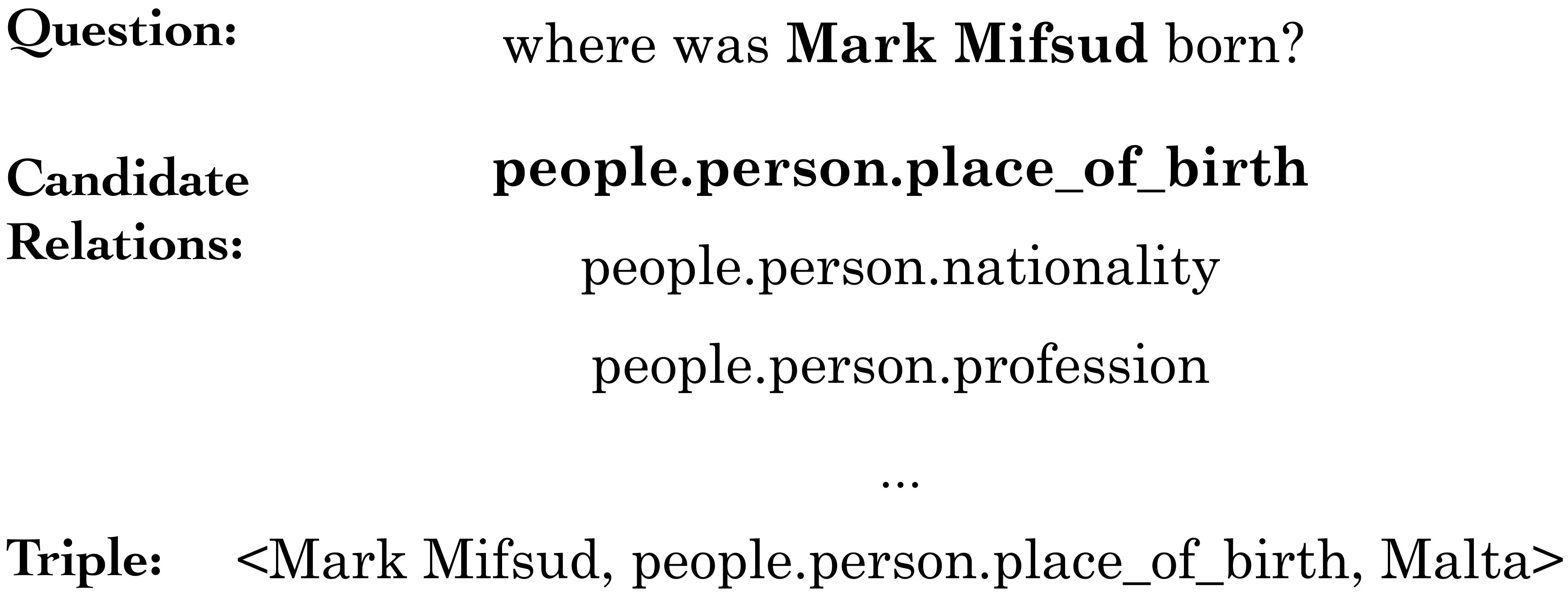}} 
\caption{ \label{fig:kbqa_example} A KBQA example. The bold words in the question are the target entity, identified in the entity linking step. The relation detection step selects the correct relation (marked with bold font) from a set of candidate relations. The answer of this question is the object entity of the triple extracted from the knowledge base.}
\end{figure}

Typical KBQA systems~\cite{yao2014information,yin2016simple,dai2016cfo,yu2017improved,hao2018pattern} can be divided into two steps: 
the entity linking step first identifies the target entity of the question, which corresponds to the subject of the triple; the relation detection step then determines the relation that the question asks from a set of candidate relations. 
After the two steps, the answer could be obtained by extracting the corresponding triple from the knowledge base (as shown in Figure \ref{fig:kbqa_example}).

Our main focus in this paper is the relation detection step, which is more challenging because it needs to consider the meaning of the whole question sentence (e.g., the pattern ``where was ... born''), as well as the meaning of the candidate relation (e.g., ``place\_of\_birth''). For comparison, the entity linking step benefits more from the matching of surface forms between the words in the question and subject entity (e.g., ``Mark Mifsud'').

In recent deep learning based relation detection approaches, each word or relation is represented by a dense vector representation, called \textit{embedding}, 
which is usually learned automatically while optimizing the relation detection objective. Then, the inference processes of these approaches are executed by neural network computations. 
Such approaches enjoy great success in common KBQA datasets, such as SimpleQuestion~\cite{bordes2015large}, achieving over 90\% accuracy in relation detection. In the words of \newcite{petrochuk2018simplequestions}, ``SimpleQuestion is nearly solved.'' However, we notice that in the common split of the SimpleQuestion dataset, 99\% of the relations in the test set also exist in the training data, which means their embeddings could be learned well during training. On the contrary, for those relations which are never seen in the training data (called \textit{unseen relations}), their embeddings have never been trained since initialization. As a result, the corresponding detection performance could be arbitrary, which is a problem that has not been carefully studied. 

We emphasize that the detection for these unseen relations is critical because it is infeasible to build training data for all the relations in a large-scale knowledge base. For example, SimpleQuestion is a large-scale human annotated dataset, which contains 108,442 natural language questions for 1,837 relations sampled from FB2M~\cite{bordes2015large}. FB2M is a subset of FreeBase~\cite{bollacker2008freebase} which have 2 million entities, 6,700 relations. A large portion of these relations can not be covered by the human-annotated dataset such as SimpleQuestion. Therefore, for building up a practical KBQA system that could answer questions based on FB2M or other large-scale knowledge bases, dealing with the unseen relations is very important and challenging. This problem could be considered as a zero-shot learning problem~\cite{palatucci2009zero} where the labels for test instances are unseen in the training dataset.

In this paper, we present a detailed study
on this zero-shot relation detection problem. Our contributions could be summarized as follows:

1. Instead of learning the relation representation barely from the training data, we employ methods to learn the representations from the whole knowledge graph which has much wider coverage. 

2. We propose a mapping mechanism, called \textit{representation adapter}, or simply \textit{adapter}, to incorporate the learned representations into the relation detection model. We start with the simple mean square error loss for the non-trivial training of the adapter and propose to incorporate adversarial and reconstruction objectives to improve the training process. 

3. We re-organize the SimpleQuestion dataset as SimpleQuestion-Balance
to evaluate the performance for seen and unseen relations, separately. 

4. We present experiments showing that our proposed method brings a great improvement to the detection of unseen relations, while still keep comparable to the state-of-the-art method for the seen relations.

\section{Representation Adapter}
\subsection{Motivation}
Representation learning of human annotated data is limited by the size and coverage of the training data. In our case, because the unseen relations and their corresponding questions do not occur in the training data, their representations cannot be properly trained, leading to poor detection performance. A possible solution for this problem is to employ a large number of unannotated data, which may be much easier to obtain, to provide better coverage. 

Usually, pre-trained representations are not directly applicable to specific tasks. 
One popular way to utilize these representations is using them as initialization.
These initialized representations are then fine-tuned on the labeled training data, with a task specific objective. However, with the above mentioned coverage issues, the representations of unseen relations will not be updated properly during fine-tuning, leading to poor test performance.

To solve this problem, we keep the representation unchanged during training, and propose a \textit{representation adapter} to bridge the gap between general purposed representations and task specific ones. We will then present the basic adapter framework, introduce the adversarial adapter and the reconstruction objective as enhancements. 

Throughout this paper, we use the following notations: let $r$ denote a single relation; $S$ and $U$ denote the set of seen and unseen relations, respectively; $e(r)$ or $e$ denote the embedding of $r$; specifically, we use $e_\text{g}$ to denote the general pre-trained embedding. 

\begin{figure}[t]
    \centering
    \includegraphics[width=0.45\textwidth]{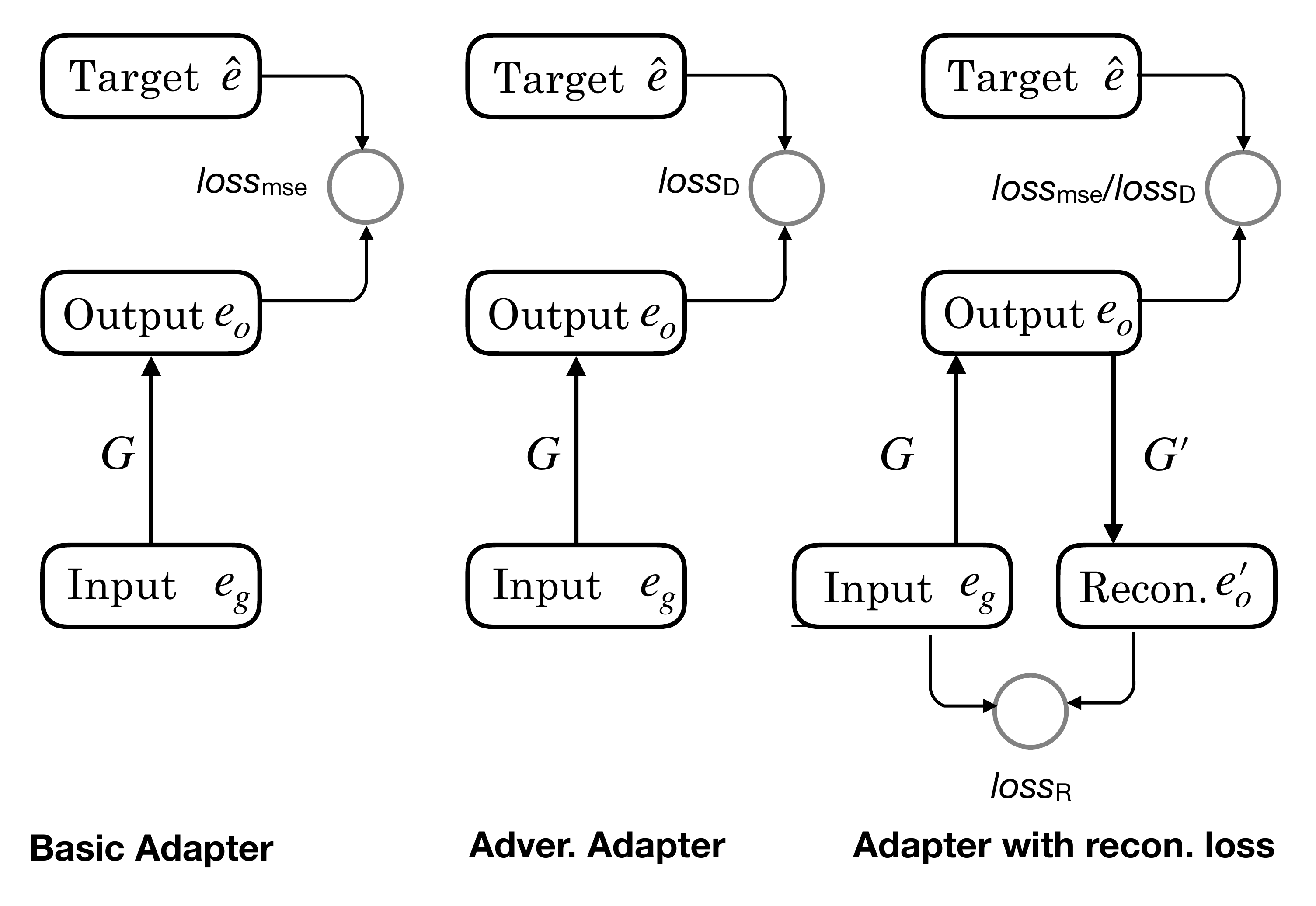}
    \caption{The structures of representation adapter.
    On the left is the basic adapter; on the middle is the adversarial adapter; on the right is the adapter with the reconstruction loss. 
    Adver. and recon. are the abbreviation of adversarial and reconstruction, respectively.
    }
    \label{fig:adapter}
\end{figure}

\subsection{Basic Adapter}

\paragraph{Pseudo Target Representations}
The basic idea is to use a neural network representation adapter to perform the mapping from the general purposed representation to the task specific one. 
The input of the adapter is the embedding learned from the knowledge base.
However, the output of the adapter is undecided, because there is no oracle representation for the relation detection task.
Therefore, we first train a traditional relation detection model similar to ~\newcite{yu2017improved}. During training, the representations for relations in the training set (seen relations) will be updated for the relation detection task. We use these representations as \textit{pseudo target representations}, denoted as $\hat{e}$, for training the adapter.

\paragraph{Linear Mapping}
Inspired by \newcite{mikolov2013exploiting}, which shows the representation space of similar languages can be transferred by a linear mapping, 
we also employ a linear mapping function $G(\cdot)$ to map the general embedding 
$e_{g}$ to the task specific (pseudo target) representation $\hat{e}$   (Figure~\ref{fig:adapter}, left).

The major difference between our adapter and an extra layer of neural network is that specific losses are designed to train the adapter, instead of implicitly learning the adapter as a part of the whole network. We train the adapter to optimize the following objective function on the seen relations:
\begin{align}\label{formula:adapter loss}
\mathcal{L}_{\text{adapter}} = \sum_{r\in{S}}loss(\hat{e}, G(e_{\text{g}})).
\end{align} 
Here the loss function could be any metric that evaluates the difference between the two representations. The most common and simple one is the mean square error loss (Equation~(\ref{formula:mse})), which we employ in our basic adapter. We will discuss other possibilities in the following sub-sections.
\begin{align}\label{formula:mse}
loss_{\text{MSE}}(\hat{e}, G(e_{\text{g}})) = ||\hat{e}- G(e_{\text{g}})||^2_2
\end{align}

\subsection{Adversarial Adapter}
The mean square error loss only measures the absolute distance between two embedding vectors. Inspired by the popular generative adversarial networks (GAN)~\cite{goodfellow2014generative,arjovsky2017wasserstein} and some previous works in unsupervised machine translation~\cite{conneau2017word,zhang2017adversarial,zhang2017earth}, we use a discriminator to provide an adversarial loss to guide the training (Figure \ref{fig:adapter}, middle). It is a different way to minimize the difference between $G(e)$ and $\hat{e}$. 

In detail, we train a discriminator, $D(\cdot)$ , to discriminate the ``real'' representation, i.e., the fine-tuned relation embedding $\hat{e}$, from the ``fake'' representation, which is the output of the adapter. The adapter $G(\cdot)$ is acting as the generator in GAN, which tries to generate a representation that is similar to the ``real'' representation. 
We use WassersteinGAN~\cite{arjovsky2017wasserstein} to train our adapter. For any relations sampled from the training set, the objective function for the discriminator ${loss}_{\text{D}}$ and generator ${loss}_{\text{G}}$ are:

\begin{align} \label{formula:adapter gan loss}
&{loss}_{\text{D}} = \mathbb{E}_{r \in S}[D(G(e_\text{g}))]-\mathbb{E}_{r \in S}[D(\hat{e})]
\end{align}
\begin{align}
&{loss}_{\text{G}} = -\mathbb{E}_{r \in S}[D(G(e_{\text{g}}))]
\end{align}
Here for $D(\cdot)$, we use a feed forward neural network without the sigmoid function of the last layer~\cite{arjovsky2017wasserstein}. 

\subsection{Reconstruction Loss}
The adapter could only learn the mapping by using the representations of seen relations, which neglects the potential large set of unseen relations. 
Here we propose to use an additional reconstruction loss to augment the adapter (Figure~\ref{fig:adapter}, right). 
More specifically, we employ a reversed adapter $G'(\cdot)$, mapping the representation $G(e)$ back to $e$.

The advantage of introducing the reversed training is two-fold. On the one hand, the reversed adapter could be trained with the representation for all the relations, both seen and unseen ones. 
On the other hand, the reversed mapping could also serve as an extra constraint for regularizing the forward mapping.

For the reversed adapter $G'(\cdot)$, We simply use a similar linear mapping function as for $G(\cdot)$, and train it with the mean square error loss:

\begin{align}\label{formula:recons loss}
{loss}_{\text{R}} = \sum_{r\in{S\cup U}}||G'(G(e_\text{g}))-e_\text{g}||^{2}_{2}
\end{align} 
Please note that, different from previous loss functions, this reconstruction loss is defined for both seen and unseen relations.


\section{Relation Detection with the Adapter}
We integrate our adapter into the state-of-the-art relation detection framework~\cite[Hierarchical Residual BiLSTM (HR-BiLSTM)]{yu2017improved}. 

\begin{figure}[t]
    \centering
    \includegraphics[width=0.45\textwidth]{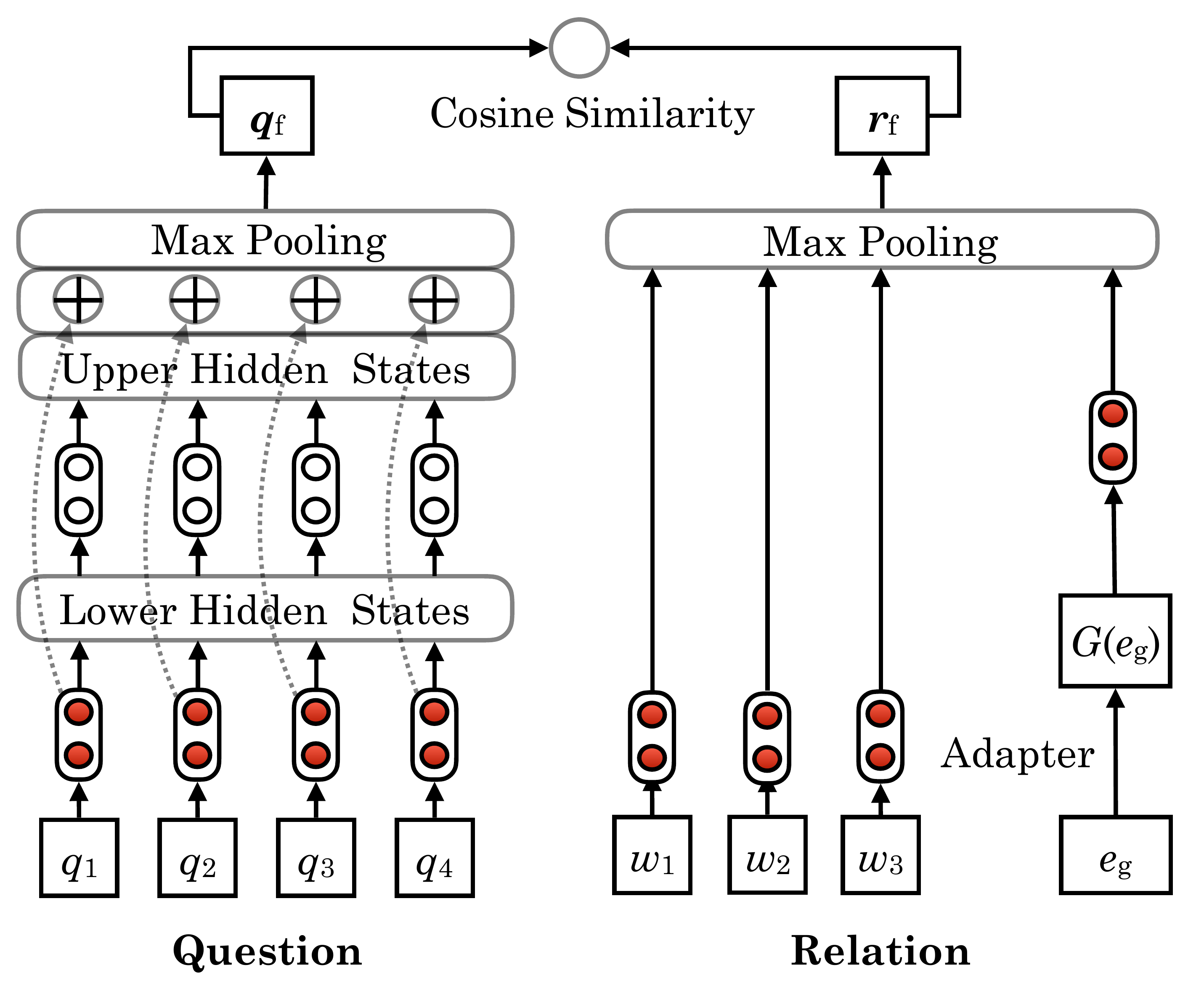}
    \caption{KBQA baseline with the adapter. Shared Bi-LSTM is marked with the same color. The adapter maps task independent representations for each relation to the task specific ones, which are fed into the relation network.}
    \label{fig:model}
\end{figure}

\paragraph{Framework}
The framework uses a question network to encode the question sentence as a vector $\mathbf{q}_\text{f}$ and a relation network to encode the relation as a vector $\mathbf{r}_\text{f}$. Both of the two networks are based on the Bi-LSTM with max-pooling operation. 
Then, the cosine similarity is introduced to compute the distance between the $\mathbf{q}_\text{f}$ and $\mathbf{r}_\text{f}$, which determines the detection result. 
Our adapter is an additional module which is used in the relation network to enhance this framework (Figure \ref{fig:model}). 

\paragraph{Adapting the Relation Representation}
The relation network proposed in \newcite{yu2017improved} has two parts for relation representations: one is at word-level and the other is at relation-level. The two parts are fed into the relation network to generate the final relation representation. 

Different from previous approaches, we employ the proposed adapter $G(\cdot)$ on the relation-level representations to solve unseen relation detection problem. 
There are several approaches to obtain the relation representations from the knowledge base into a universal space~\cite{bordes2013translating,wang2014knowledge,han2018neural}. In practice, we use the JointNRE embedding \cite{han2018neural}, because its word and relation representations are in the same space.

\paragraph{Training} 
Following~\newcite{yu2017improved}, the relation detection model is trained by the hinge loss \cite{bengio2003neural} which tries to separate the score of each negative relation from the positive relation by a margin:
\begin{equation} \label{formula:score_loss}
\mathcal{L}_{\text{rd}} = \sum \max (0,\gamma -
s(\mathbf{q}_\text{f},\mathbf{r}_\text{f}^{+})+s(\mathbf{q}_\text{f},\mathbf{r}_\text{f}^{-})),
\end{equation}
where $\gamma$ is the margin; $\mathbf{r}_\text{f}^{+}$ is the positive relation from the annotated training data; $\mathbf{r}_\text{f}^{-}$ is the relation negative sampled from the rest relations; $s(\cdot,\cdot)$ is the cosine distance between $\mathbf{q}_\text{f}$ and $\mathbf{r}_\text{f}$.

The basic relation detection model is pre-trained to get the pseudo target representations. Then, the adapter is incorporated into the training process, and jointly optimized with the relation detection model. 
For the adversarial adapter, the generator and the discriminator are trained alternatively following the common practice.  

\section{SimpleQuestion-Balance (SQB)} \label{section:unseen_sq}


As mentioned before, SimpleQuestion (SQ) is a large-scale KBQA dataset. Each sample in SQ includes a human annotated question and the corresponding knowledge triple. However, the distribution of the relations in the test set is unbalanced. Most of the relations in the test set have been seen in the training data. To better evaluate the performance of unseen relation detection, we re-organize the SQ dataset to balance the number of seen and unseen relations in development and test sets, and the new dataset is denoted as \textit{SimpleQuestion-Balance (SQB)}. 

\begin{table}[ht]
\footnotesize
\begin{center}
\begin{tabular}{l||c|c}
\toprule
{Datasets} & SQ & SQB \\
\hline
{Train} & 75,910 & $\text{75,819}$ \\
\hline
{Dev-seen} &10,774&$\text{5,383}$ \\
 {Dev-unseen}&71&\bm{$\textbf{5,758}$} \\
 \hline
 {Test-seen}& 21,526&$\text{10,766}$ \\
 {Test-unseen}& 161&\bm{$\textbf{10,717}$} \\ 
 \bottomrule

\end{tabular}
\end{center}
\caption{\label{table:detail_dataset}The number of instances in each subset from SimpleQuestion (SQ) and SimpleQuestion-Balance (SQB) datasets. Dev-seen and Dev-unseen are seen and unseen part of development set; Test-seen and Test-unseen are seen and unseen part of test set, respectively.}
\end{table}

The re-organization is performed by randomly shuffle and split into 5 sets, i.e. Train, Dev-seen, Den-unseen, Test-seen and Test-unseen, while checking the overlapping of relations and the percentage of seen/unseen samples in each set. We require the sizes of the training, development and test sets are similar to SQ.

The details of the resulting SQB and SQ are shown in Table \ref{table:detail_dataset}. 
The SQ dataset only have 0.65\% (71 / 10845) and 0.74\% (161 / 21687) of the unseen samples in the dev set (Dev-unseen) and test set (Test-unseen), respectively. 

\section{Experiment} \label{section:exp}

\begin{table*}[!ht]
\footnotesize
\begin{center}

\begin{tabular}{c|l||ccc}
\toprule
 \multirow{2}{*}{\#}  & \multirow{2}{*}{Model}      & \multicolumn{3}{c}{Micro / Macro Average Accuracy  on SQB  (\%)} \\ 
   &   & {Test-seen}   & {Test-unseen}  & All   \\ \hline
 
1&HR-BiLSTM      & \textbf{93.5$\pm$0.6}  / 84.7$\pm$1.4 & 33.0$\pm$5.7 / 49.3$\pm$1.7 & 63.3$\pm$3.6 / 71.2$\pm$1.3   \\

2&\ \ + no fine-tune     & 93.4$\pm$0.7  / 83.8$\pm$0.7 & 57.8$\pm$9.8 / 60.8$\pm$2.0 & 75.6$\pm$5.0 / 75.0$\pm$0.6 \\  

\hline
3&\ \ + no fine-tune + mapping     & 93.3$\pm$0.7 / 84.0$\pm$1.6 & 52.0$\pm$7.2 / 60.6$\pm$2.1 & 72.7$\pm$3.8  / 75.1$\pm$1.3     \\ \hline

4&\ \ + Basic-Adapter        & 92.8$\pm$0.7 / 84.1$\pm$1.2 & 76.0$\pm$7.5$^\dag$ / 69.5$\pm$2.0$^\dag$ & 84.5$\pm$3.5 / 78.5$\pm$1.3   \\

5&\ \ \ \ + reconstruction     & 93.0$\pm$0.5 / 84.4$\pm$0.8 & 76.1$\pm$7.0$^\dag$ / 70.7$\pm$1.8$^\dag$ &  84.6$\pm$3.3  / 79.2$\pm$0.8    \\ \hline

6&\ \ + Adversarial-Adapter      & 92.6$\pm$0.9 / \textbf{86.4$\pm$1.4} & 77.1$\pm$7.1$^\dag$   / \textbf{73.2$\pm$2.1$^\dag$} & \textbf{84.9$\pm$3.2} / \textbf{ 81.4$\pm$1.4}   \\ 

7&\ \ \ \ + reconstruction [Final]     & 92.4$\pm$0.8 / {86.1$\pm$0.7} & \textbf{77.3$\pm$7.6$^\dag$}  / 73.0$\pm$1.7$^\dag$ &  \textbf{84.9$\pm$3.5}  / {81.1$\pm$0.8}  \\  


\bottomrule 

\end{tabular}

\end{center}
\caption{\label{table:result} The micro average accuracy and macro average accuracy of relation detection on the SQB dataset.
``$^\dag$'' indicates statistically significant difference ($p<0.01$) from the HR-BiLSTM.}
\end{table*}

\subsection{Settings}
\paragraph{Implementation Details} We use RMProp~\cite{tieleman2012lecture} as the optimization strategy to train the proposed adapter. The learning rate is set as $10^{-4}$. We set the batch size as 256. Following \newcite{arjovsky2017wasserstein}, we clip the parameters of discriminator into $[-c,c]$, where $c$ is 0.1. Dropout rate is set as 0.2 to regularize the adapter. 

The baseline relation detection model is almost same as~\newcite{yu2017improved}, except that the word embedding and relation embedding of our model are pre-trained by JointNRE~\cite{han2018neural} on FB2M and Wikipedia
, with the default settings reported in the~\newcite{han2018neural}. The embeddings are fine-tuned with the model.

More specifically, the dimension of relation representation is 300. The dimension for the hidden state of Bi-LSTM is set to 256. Parameters in the neural models are initialized using a uniform sampling. The number of negative sampled relations is 256. The $\gamma$ in hinge loss (Equation (\ref{formula:score_loss})) is set to 0.1. 



\paragraph{Evaluation} 
To evaluate the performance of relation detection, we assume that the results of entity linking are correct. 
Two metrics are employed. \textit{Micro average accuracy}~\cite{tsoumakas2009mining} is the average accuracy of all samples, which is the metric used in previous work. \textit{Macro average accuracy}~\cite{sebastiani2002machine,Manning2008IIR1394399,tsoumakas2009mining} is the average accuracy of the relations. 

Please note that because different relations may correspond to the different number of samples in the test set, the micro average accuracy may be affected by the distribution of unseen relations in the test set. In this case, the macro average accuracy will serve as an alternative indicator. 

We report the average and standard deviation (std) of 10-folds cross validation to avoid contingency.

\subsection{Main Results}
Main results for baseline and the proposed model with the different settings are listed in Table \ref{table:result}. The detailed comparison is as follows:

\paragraph{Baseline} 
The baseline HR-BiLSTM (line 1) shows the best performance on Test-seen, but the performance is much worse on Test-unseen. For comparison, training the model without fine-tuning (line 2) achieves much better results on Test-unseen, demonstrating our motivation that the embeddings are the reason for the weak performance on unseen relations, and fine-tuning makes them worse.

\paragraph{Using Adapters} 
Line 3 shows the results of adding an extra mapping layer of neural networks between the pretrained embedding and the relation detection networks, without any loss. Although ideally, it is possible to learn the mapping implicitly with the training, in practice, this does not lead to a better result (line 3 v.s. line 2).

While keeping similar performance on the Test-seen with the HR-BiLSTM, all the models using the representation adapter achieve great improvement on the Test-unseen set. With the simplest form of adapter (line 4), the accuracy on Test-unseen improves to 76.0\% / 69.5\%. It shows that our model can predict unseen relation with better accuracy. 

Using adversarial adapter (line 6) can further improve the performance on the Test-unseen in both micro and macro average scores. 

\paragraph{Using Reconstruction Loss}
Adding reconstruction loss to basic adapter can also improve the performance (line 5 v.s. line 4) slightly. The similar improvement is obtained for the adversarial adapter in micro average accuracy (line 7 v.s. line 6). 

Finally, using all the techniques together (line 7) gets the score of 77.3\% / 73.0\% on Test-unseen, and 84.9\% / 81.1\% on the union of Test-seen and Test-unseen in micro/macro average accuracy, respectively. We mainly use this model as our final model for further comparison and analysis.


We notice that the results of our model on Test-seen are slightly lower than that of HR-BiLSTM. It is because we use the mapped representations for the seen relations instead of the directly fine-tuned representations. This dropping is negligible compared with the improvement in the unseen relations.

\paragraph{Integration to the KBQA}
To confirm the influence of unseen relation detection for the entire KBQA, we integrate our relation detection model into a prototype KBQA framework. During the entity linking step, we use FocusPrune \cite{dai2016cfo} to get the mention of questions. Then, the candidate mentions are linked to the entities in the knowledge base. Because the FreeBase API was deprecated \footnote{\url{https://developers.google.com/freebase/}}, we restrict the entity linking to an exact match for simplicity. The candidate relations are the set of relations linked with candidate subjects. We evaluate the KBQA results using the micro average accuracy introduced in \newcite{bordes2015large}, which considers the prediction as correct if both the subject and relation are correct.

As shown in Table \ref{table:kbqa acc}, the proposed adapter method can improve KBQA from 48.5\% to 63.7\%. Comparing with the result of relation detection, we find that the boost of relation detection could indeed lead to the improvement of a KBQA system.

\begin{table}[t]
\footnotesize
\begin{center}
\begin{tabular}{l||c}
\toprule
Model                  & Accuracy (\%)\\ \hline
HR-BiLSTM            & 48.5$\pm$3.3        \\ 
\ \ + no fine-tune & 56.4$\pm$3.4 \\ 
\hline
Final  & \textbf{63.7$\pm$3.2} \\ \bottomrule
\end{tabular}
\end{center}
\caption{\label{table:kbqa acc} The micro average accuracy of the whole KBQA system with different relation detection models.}
\end{table}

\begin{table}[t]
\footnotesize
\begin{center}
\begin{tabular}{l||c}
\toprule
Model                  &  Seen Rate $\downarrow$ (\%)\\ \hline
HR-BiLSTM        & 47.2$\pm$2.0  \\ 
\ \ + no fine-tune & 34.8$\pm$2.3 \\ \hline
Final & \textbf{21.2$\pm$1.7}  \\ \bottomrule
\end{tabular}
\end{center}
\caption{\label{table:error anaysis} Seen relation prediction rate in the Test-unseen set. We calculate the macro average of this rate.}
\end{table}    

\subsection{Analysis}

\paragraph{Seen Relation Bias} 
We use macro-average to calculate the percentage of instances whose relations are wrongly predicted to be a seen relation on Test-unseen. We call this indicator the \textit{seen rate}, the lower the better. Because the seen relations are better learned after fine-tuning, while the representations for unseen relations are not updated well. So the relation detection model may have a strong trend to select those seen relations as the answer. 
The result in Table \ref{table:error anaysis} shows that our adapter makes the trend of choosing seen relation weaker, which helps to promote a fair choice between seen and unseen relations.


\begin{figure}[t]
\centering
{
\includegraphics[width=7cm]{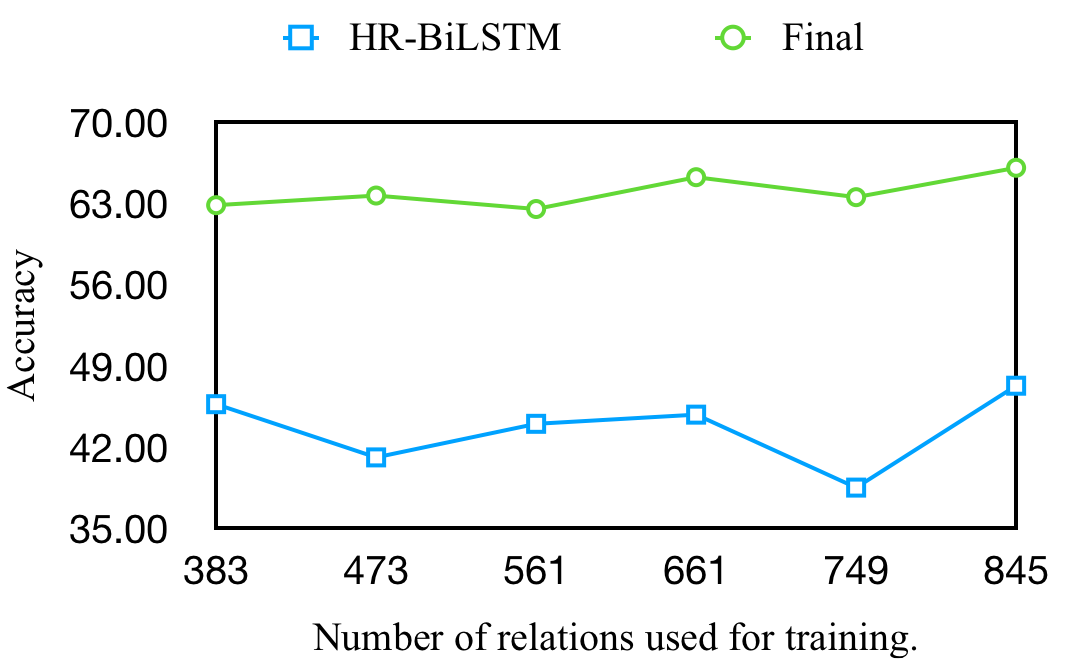}} 
\caption{ \label{fig:diff_relation_size} Macro average accuracy for different relation size in the training set.}
\end{figure}

\paragraph{Influence of Number of Relations for Training} We discuss the influence of the number of relations in the training set for our adapter. Our adapter are trained mainly by the seen relations, because we can get pseudo target representation for these relations. In this experiment, we random sample 60,000 samples from the training set to perform the training, and plot the accuracy against the different number of relations for training. We report the macro average accuracy on Test-unseen. 
\begin{figure}[ht]
\centering
\subfloat[JointNRE]{
 \label{fig:jointnre_vis}
\includegraphics[scale=.15]{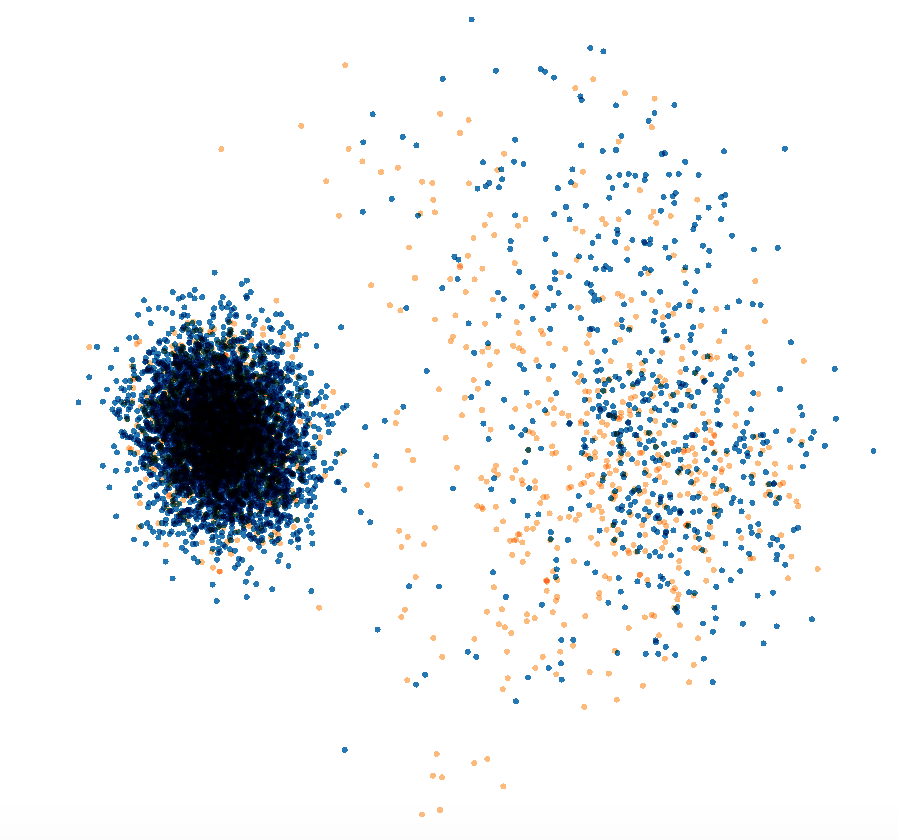}
}
\subfloat[HR-BiLSTM]{
 \label{fig:baseline_vis}
\includegraphics[scale=.15]{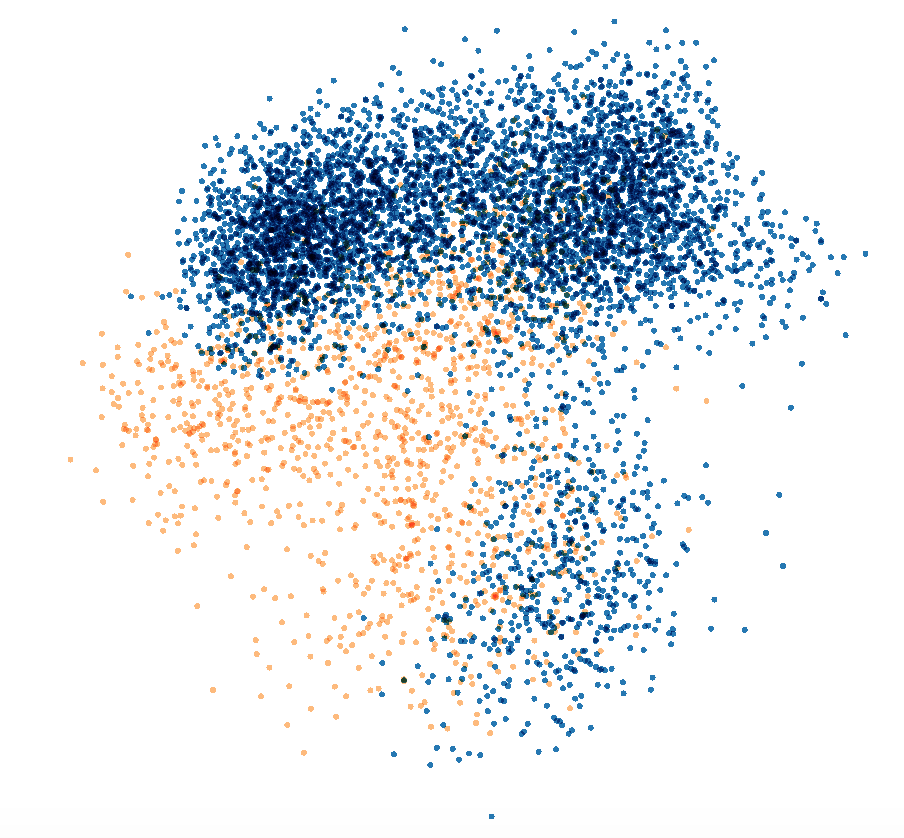}} 
\subfloat[Final]{
 \label{fig:adapter_vis}
\includegraphics[scale=.15]{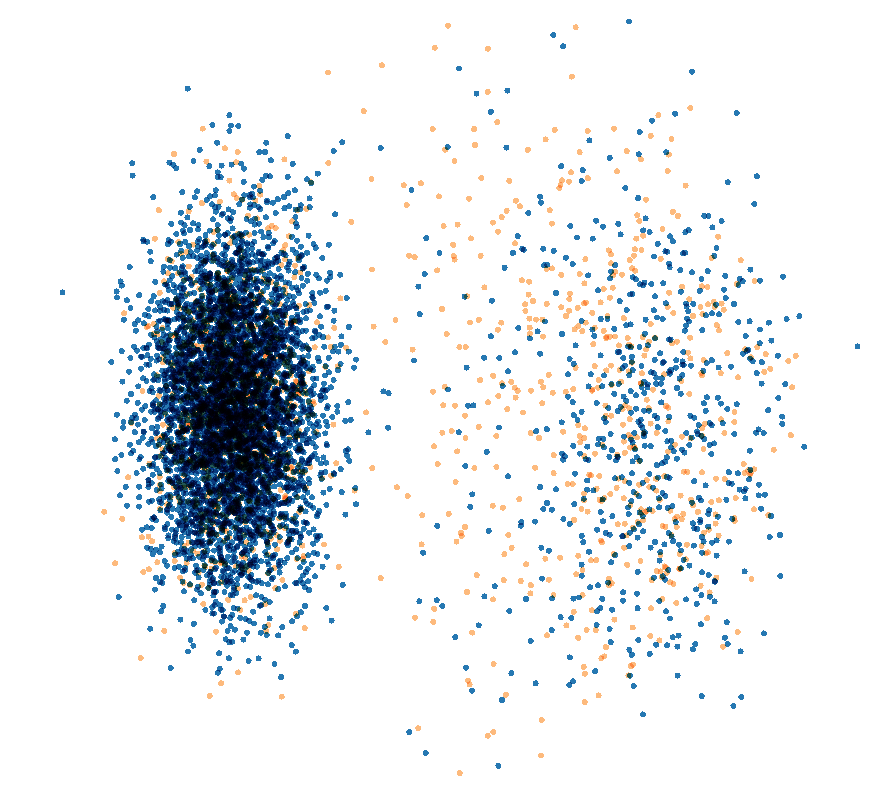}} \\
\subfloat[JointNRE*]{
 \label{fig:jointnre_star}
\includegraphics[scale=.15]{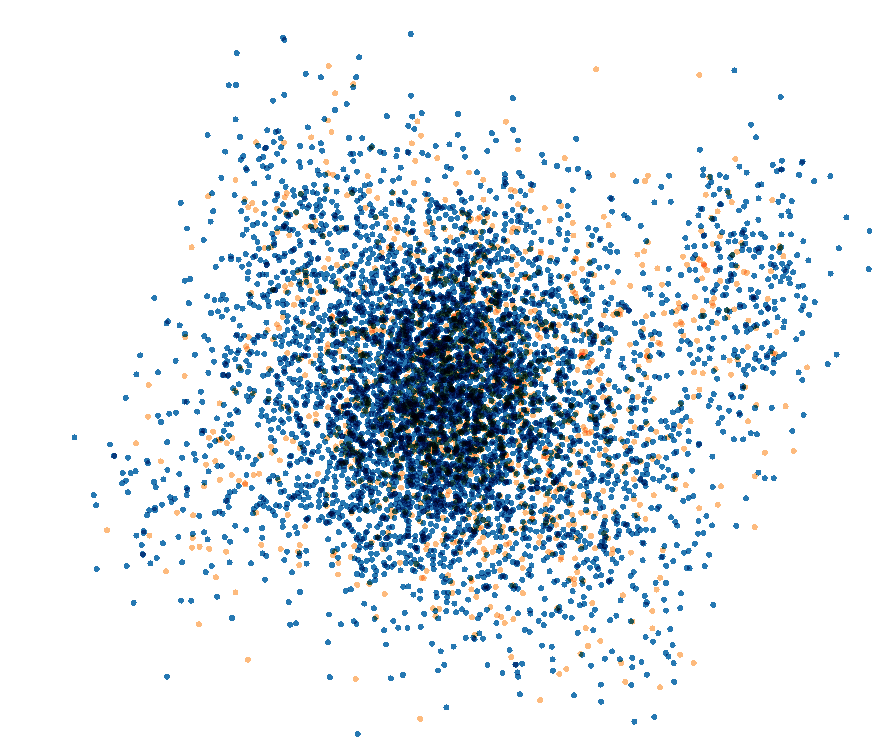}}
\subfloat[HR-BiLSTM*]{
 \label{fig:baseline_star}
\includegraphics[scale=.15]{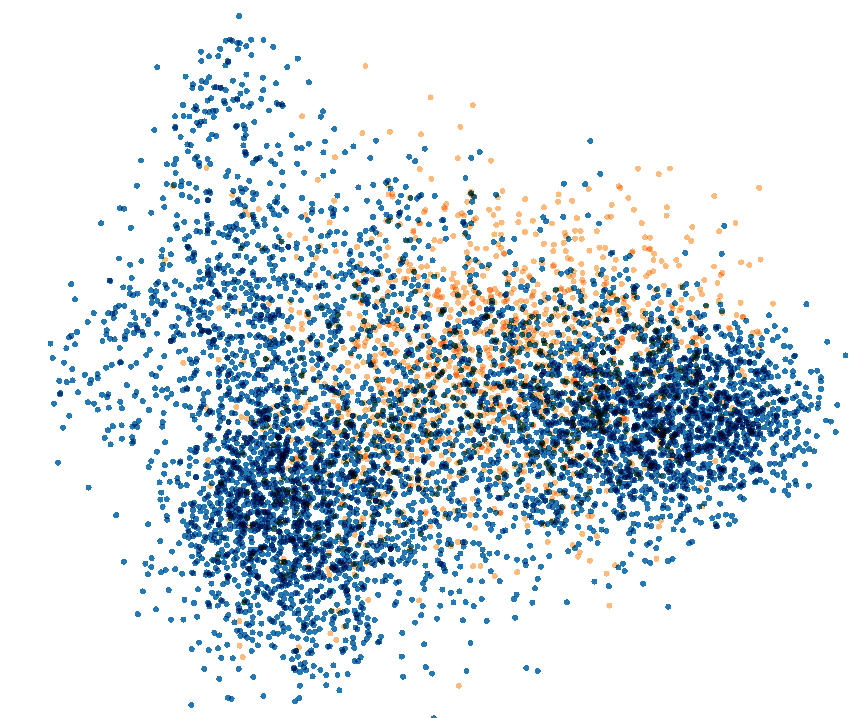}}
\subfloat[Final*]{
 \label{fig:fianl_star}
\includegraphics[scale=.15]{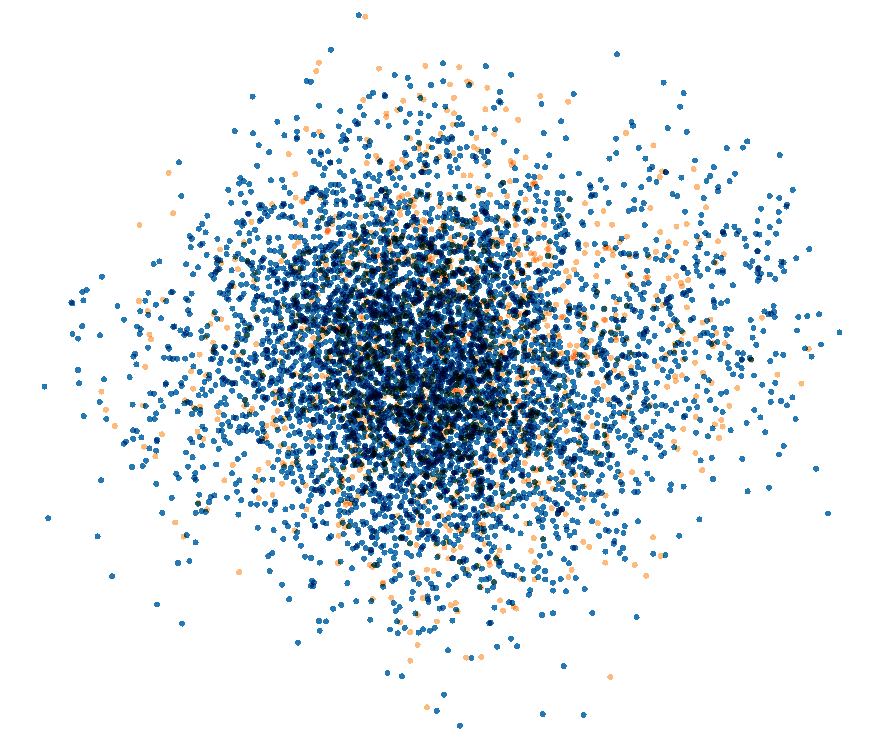}}

\label{fig:vis_rep}
\caption{Relation Representation Visualization of different models. The yellow (light) point represent the seen relation, and the blue (dark) point represent the unseen relation.}
\end{figure}

As shown in Figure \ref{fig:diff_relation_size}, with different number of relations, our model still perform better than HR-BiLSTM. Note that, our adapter can beat HR-BiLSTM with even a smaller number of seen relations.  When there are more relations for training, the performance will be improved as expected. 



\paragraph{Relation Representation Analysis} 
We visualize the relation representation in JointNRE, HR-BiLSTM and the output representation of our final adapter by principal component analysis (PCA) with the help of TensorBoard. We use the yellow (light) point represents the seen relation, and the blue (dark) point represents the unseen relation. 

As shown in Figure~\ref{fig:jointnre_vis}), the JointNRE representation is pre-trained by the interaction between knowledge graph and text. Because without knowing the relation detection tasks, seen and unseen relations are randomly distributed.  \footnote{We also notice that there is a big cluster of relations on the left hand side. This is presumably the set of less updated relations in the training of JointNRE, due to lack of correspondence with the text data. This cluster does not affect our main observation with adapter training.}


After training with HR-BiLSTM (Figure~\ref{fig:baseline_vis}), the seen and unseen relations are easily separated, because the training objective is to discriminate the seen relations from the other relations for the corresponding question. Although the embeddings of unseen relations are also updated due to negative sampling, they are never updated towards their correct position in the embedding space. As a result, the relation detection accuracy for the unseen relations is poor.

The training of our final model uses the adapter to fit the training data, instead of directly updating the embeddings. Despite the comparable performance on seen relations, the distribution of seen and unseen relations (Figure~\ref{fig:adapter_vis}) is much similar to the original JointNRE, which is the core reason for its ability to obtain better results on unseen relations. 

\paragraph{Adapting JointNRE}
Interestingly, we notice that JointNRE is to train the embedding of relations with a corpus of text that may not cover all the relations, which is also a process that needs the adapter.
As a simple solution, we use a similar adapter to adapt the representation from TransE~\footnote{\url{https://github.com/thunlp/Fast-TransX}}~\cite{lin2015learning} to the training of JointNRE. With the resulting relation embedding, denoted as JointNRE*, we train the baseline and final relation detection models, denoted as HR-BiLSTM* and Final*, respectively.

We visualize the relation representation in these models again. Clearly, the distribution of seen and unseen relations in JointNRE* (Figure \ref{fig:jointnre_star}) looks more reasonable than before. This distribution is interrupted by fine-tuning process of HR-BiLSTM* (Figure \ref{fig:baseline_star}), while is retained by our adapter model (Figure \ref{fig:fianl_star}).

\begin{table}[t]
\footnotesize
\begin{center}
\begin{tabular}{l||c}
\toprule
Model                  & Accuracy \\ \hline

 Final & 77.3$\pm$7.6	/ 73.0$\pm$1.7    \\ \hline
Final*  & \textbf{77.5$\pm$6.0} / 72.4$\pm$1.8 \\ \bottomrule
\end{tabular}
\end{center}
\caption{\label{table:diff_rep} Results on Test-unseen with and without the adapter in training JointNRE.}
\end{table}



Furthermore, as shown in Table \ref{table:diff_rep}, using JointNRE* can further improve the unseen relation detection performance (77.5\% v.s. 77.3\%). This provides further evidence of the importance of representations for unseen relations. 

\begin{table}[t]
\footnotesize
\begin{center}
\begin{tabular}{l||l}
\toprule


 Question 1&  \tabincell{l}{who produced recording \textit{Twenty One}}     \\ \hline
\tabincell{l}{Candidate \\Relations\\} & 
\tabincell{l}{ \textbf{music.recording.producer}\\ music.recording.artist
}
 \\  \hline
HR-BiLSTM & music.recording.artist  \\  \hline
Final &\textbf{music.recording.producer} \\ \hline  \hline
  Question 2&  \tabincell{l}{what is \textit{Tetsuo Ichikawa}'s profession}     \\  \hline
\tabincell{l}{Candidate \\Relations\\} & 
\tabincell{l}{people.person.gender \\ \textbf{people.person.profession}\\
}
 \\  \hline
HR-BiLSTM & \textbf{people.person.profession}  \\  \hline
Final &\textbf{people.person.profession} \\  
\hline \hline

 Question 3&  \tabincell{l}{which village is in \textit{Arenac county} ?}     \\ \hline
\tabincell{l}{Candidate \\Relations\\} & 
\tabincell{l}{location.us\_county.hud\_county\_place \\ \textbf{location.location.contains}\\
}
 \\ \hline
HR-BiLSTM & location.us\_county.hud\_county\_place  \\  \hline
Final & location.us\_county.hud\_county\_place \\  
\bottomrule
\end{tabular}
\end{center}
\caption{\label{table:case study} Case studies for relation detection using different models. For each question, the gold relation is marked with bold font;  the gold target entity of the question is marked with italic font. The models and notations are the same as in Table \ref{table:result}.
}
\end{table}

\paragraph{Case Study}
In the first case of Table \ref{table:case study}, \textit{Twenty One} is the subject of question. ``music.recording.producer'' is the gold relation, but it is an unseen relation. The baseline model predicts ``music.recording.artist'' because this relation is seen and perhaps relevant in the training set. A dig into the set of relations shown that there is a seen relation, ``music.recording.engineer'', which happens to be the closest relation in the mapped representation to the gold relation. It is possible that the knowledge graph embedding 
is able to capture the relatedness between the two relations. 

In the second case, although the gold relation ``people.person.profession'' is unseen, both baseline and our model predict the correct answer because of strong lexical evidences: ``profession''. 

In the last case, both the gold relation and predict error relation are unseen relation. ``Hud\_county\_place'' refers to the name of a town in a county, but ``location.location.contains'' has a broader meaning. When asked about ``village'', ``location.location.contains'' is more appropriate. This case shows that our model still can not process the minor semantic difference between word.  We will leave it for future work.

\section{Related Work}

\paragraph{Relation Detection in KBQA}
\newcite{yu2017improved} first noticed the zero-shot problem in KBQA relation detection. They split relation into word sequences and use it as a part of the relation representation. In this paper, we push this line further and present the first in-depth discussion about this zero-shot problem. We propose the first relation-level solution and present a re-organized dataset for evaluation as well. 
    
\paragraph{Embedding Mapping}
Our main idea of embedding mapping is inspired by previous work about learning the mapping of bilingual word embedding. \newcite{mikolov2013exploiting} observed the linear relation of bilingual word embedding, and used a small starting dictionary to learn this mapping. 
\newcite{zhang2017adversarial} use Generative Adversarial Nets~\cite{goodfellow2014generative} to  learn the mapping of bilingual word embedding in an unsupervised manner.
Different from this work which maps words in different languages, we perform mappings between representations generated from heterogeneous data, i.e., knowledge base and question-triple pairs. 

\paragraph{Zero-Shot Learning} Zero-shot learning has been studied in the area of natural language process.
\newcite{hamaguchi2017knowledge} use a neighborhood knowledge graph as a bridge between out of knowledge base entities to train the knowledge graph. \newcite{levy2017zero} connect nature language question with relation query to tackle zero shot relation extraction problem. \newcite{elsahar2018zero} extend the copy actions~\cite{luong2014addressing} to solve the rare words problem in text generation.
Some attempts have been made to build machine translation systems for language pairs without direct parallel data, where they relying on one or more other languages as the pivot \cite{firat2016zero, ha2016toward, chen2017teacher}.
In this paper, we use knowledge graph embedding as a bridge between seen and unseen relations, which shares the same spirit with previous work. However, less study has been done in relation detection. 

\section{Conclusion}
In this paper, we discuss unseen relation detection in KBQA, where the main problem lies in the learning of representations. We re-organize the SimpleQuestion dataset as SimpleQuestion-Balance to reveal and evaluate the problem, and propose an adapter which significantly improves the results.

We emphasize that for any other tasks which contain a large number of unseen samples, training, fine-tuning the model according to the performance on the seen samples alone is not fair. 
Similar problems may exist in other NLP tasks, which will be interesting to investigate in the future.

\section*{Acknowledgement}
We would like to thank the anonymous reviewers for their insightful comments. Shujian Huang is the corresponding author. This work is supported by the National Science Foundation of China (No. 61772261), the Jiangsu Provincial Research Foundation for Basic Research (No. BK20170074). Part of this research is supported by the Huawei Innovation Research Program (HO2018085291).

\bibliography{acl2019}
\bibliographystyle{acl_natbib}

\end{document}